\documentclass[letterpaper, 10 pt, conference]{ieeeconf}

\IEEEoverridecommandlockouts

\usepackage{graphicx}
\usepackage{glossaries}
\usepackage{amsmath}
\usepackage{caption}
\usepackage{subcaption}
\usepackage{xcolor}
\usepackage{booktabs}
\usepackage{textcomp}
\usepackage{gensymb}
\usepackage[normalem]{ulem}
\usepackage{xspace}
\usepackage{hyperref}
\usepackage{soul}
\usepackage{float}
\usepackage{tabularray} 
\usepackage{cite}

\usepackage{pgfplots}
\usepackage{multirow}
\usepgfplotslibrary{colorbrewer}
\pgfplotsset{compat = 1.15, cycle list/Set1-8} 
\usetikzlibrary{pgfplots.statistics, pgfplots.colorbrewer} 
\usepackage{pgfplotstable}
\usepackage{filecontents}

\hypersetup{
    colorlinks,
    linkcolor={red!50!black},
    citecolor={blue!50!black},
    urlcolor={blue!80!black}
}

\definecolor{warningcolor}{RGB}{204,51,20}
\definecolor{fernandoblue}{RGB}{65,105,225}

\newacronym{cis}{CIS}{CMOS-based image sensor}
\newacronym{imo}{IMO}{independently moving object}
\newacronym{hd}{HD}{high-definition}
\newacronym{als}{ALS}{Airborne Laser Scanning}
\newacronym{dbh}{DBH}{diameter at breast height}
\newacronym{mls}{MLS}{Mobile Laser Scanning}
\newacronym{tls}{TLS}{Terrestrial Laser Scanning}
\newacronym{uls}{ULS}{UAV Laser Scanning}
\newacronym{uav}{UAV}{Unmanned Aerial Vehicle}

\newcommand{\datasetname}{{TreeScope}\xspace}
\newcommand{\sensortower}{{Sensor platform}\xspace}
\newcommand{\STAB}[1]{\begin{tabular}{@{}c@{}}#1\end{tabular}}

\title{\datasetname: An Agricultural Robotics Dataset for LiDAR-Based Mapping of Trees in Forests and Orchards}
\author{
Derek Cheng$^\dagger$, Fernando Cladera$^\dagger$, Ankit Prabhu$^\dagger$, Xu Liu$^\dagger$, Alan Zhu$^\dagger$, \\ P. Corey Green$^\ddagger$, Reza Ehsani$^{\dagger\dagger}$, Pratik Chaudhari$^\dagger$, and Vijay Kumar$^\dagger$%
\thanks{We gratefully acknowledge the support of the IoT4Ag ERC funded by the National Science Foundation (NSF) under NSF Cooperative Agreement Number EEC-1941529, NIFA grant 2022-67021-36856, NSF grant CCR-2112665, and C-BRIC, a Semiconductor Research Corporation Joint University Microelectronics Program cosponsored by DARPA.}
\thanks{
$^\dagger$GRASP Laboratory, University of Pennsylvania, Philadelphia, PA, 19104, USA 
    {\tt\small \{derekch, fclad, praankit, liuxu, alzhu, pratikac, kumar\}@seas.upenn.edu}}%
\thanks{
$^\ddagger$Virginia Polytechnic Institute and State University, Forest Resources and Environmental Conservation, Blacksburg, VA 24061, USA 
    {\tt\small pcgreen7@vt.edu}}%
\thanks{
$^{\dagger\dagger}$Department of Mechanical Engineering, University of California, Merced, CA, 95343, USA 
    {\tt\small rehsani@ucmerced.edu}}%
    }

\date{September 2022}

\begin{document}

\maketitle
\thispagestyle{empty}
\pagestyle{empty}

\begin{abstract}
Data collection for forestry, timber, and agriculture relies on manual techniques which are labor-intensive and time-consuming. 
We seek to demonstrate that robotics offers improvements over these techniques and can accelerate agricultural research, beginning with semantic segmentation and diameter estimation of trees in forests and orchards. 
We present \datasetname v1.0, the first robotics dataset for precision agriculture and forestry addressing the counting and mapping of trees in forestry and orchards.
\datasetname provides LiDAR data from agricultural environments collected with robotics platforms, such as UAV and mobile robot platforms carried by vehicles and human operators. 
In the first release of this dataset, we provide ground-truth data with over 1,800 manually annotated semantic labels for tree stems and field-measured tree diameters. 
We share benchmark scripts for these tasks that researchers may use to evaluate the accuracy of their algorithms.
Finally, we run our open-source diameter estimation and off-the-shelf semantic segmentation algorithms and share our baseline results.
\end{abstract}
The dataset can be found at \url{https://treescope.org}, and the data pre-processing and benchmark code is available at \url{https://github.com/KumarRobotics/treescope}.

\section{Introduction}
Counting and mapping trees is a critical task for forestry and crop management. In particular, diameters of trees correlate with the wood biomass ~\cite{brown1997estimating} and are used to assess forest growth, determine carbon sequestration potential, evaluate fire risk management, and quantify timber yield \cite{Calders2022Carbon}. 
Traditional methods for measuring tree diameters rely on using manual tools such as calipers and measuring tapes ~\cite{Krisanski2021Sensor}, which are time-consuming, labor-intensive, and prone to errors for untrained users.

Robotics provides a unique opportunity to overcome the limitations of traditional methods and revolutionize forestry operations, measuring tree diameters and forest biomass in a non-destructive and efficient manner. \cite{White2016Remote}
Additionally, robotic measurements can be performed at a much larger scale and with higher precision, improving the accuracy of forest inventories and leading to more informed forest management decisions~\cite{liu2022challenges, liu2021large}.

\begin{figure}[t]
  \centering
  \subfloat{\includegraphics[width=0.8\linewidth]{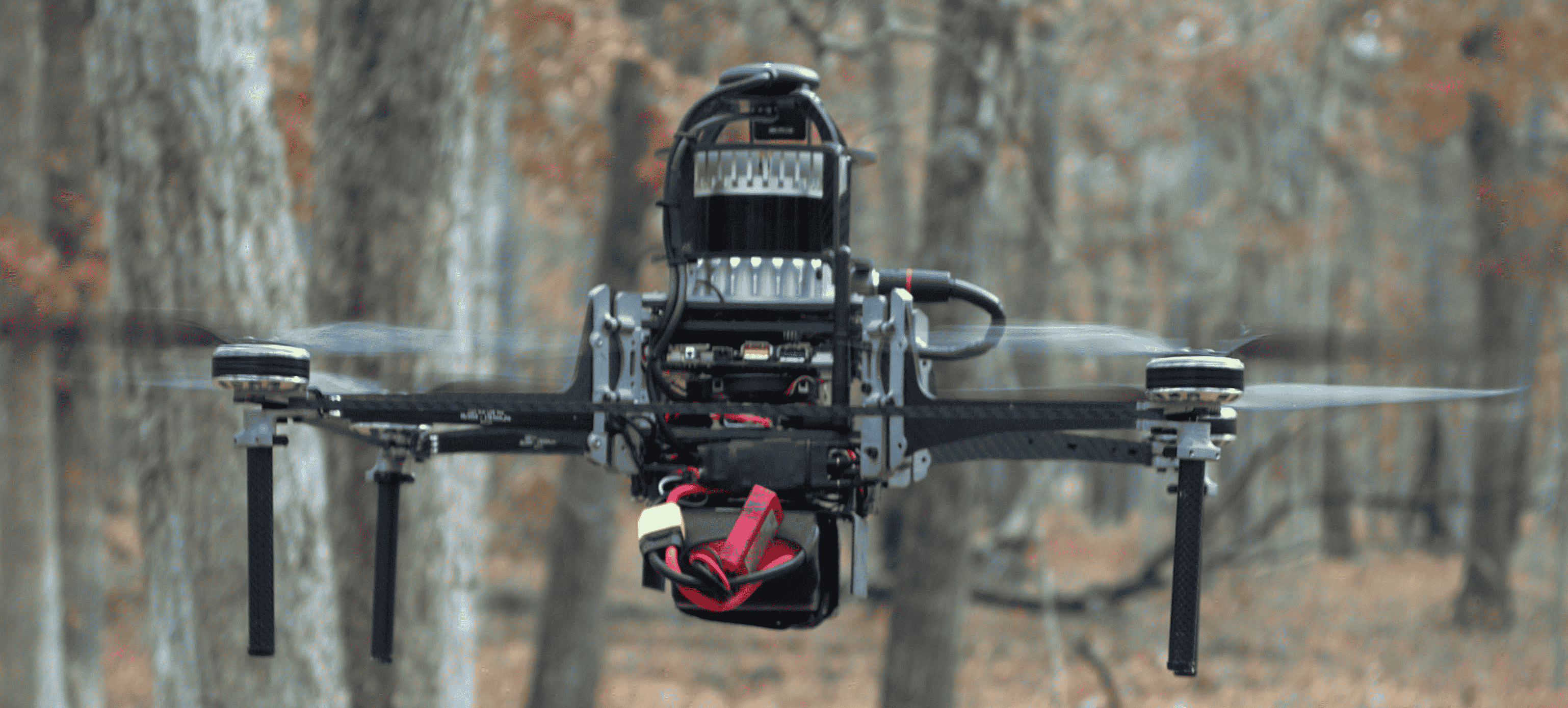}} 
  \vspace{2pt}
  \subfloat{\includegraphics[width=0.8\linewidth]{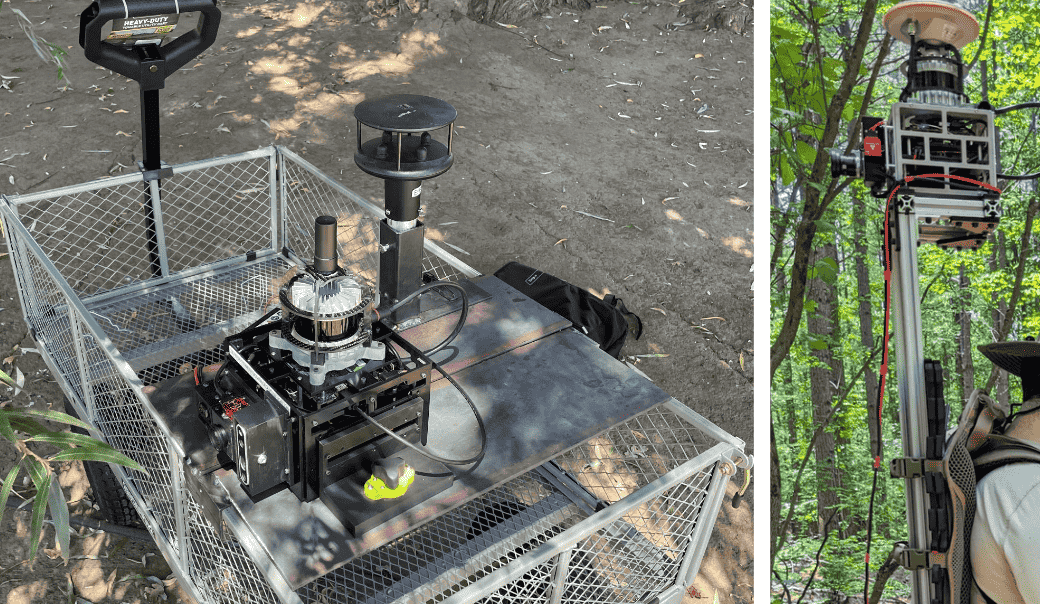}}
  \vspace{2pt}
  \subfloat{\includegraphics[width=0.8\linewidth]{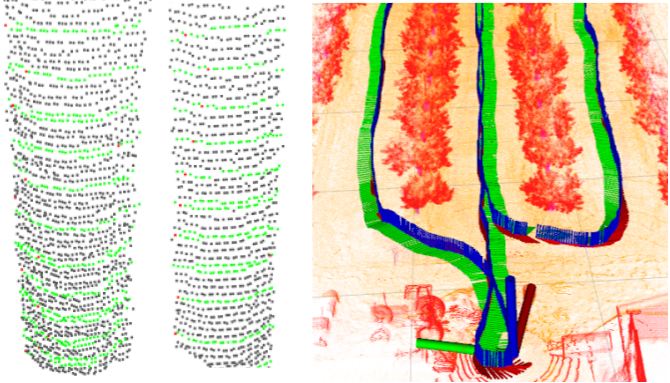}    }
  \caption{Overview of the methods for \datasetname data collection \ul{Top:} Falcon 4 UAV \cite{liu2021large} flying in under-canopy dense forests. \ul{Middle:} Our sensor platform mounted onto a small cart and human-carried backpack. \ul{Bottom left}: Forestry tree instance segmentation for diameter estimation. \ul{Bottom right}: UAV flight in a pistachio orchard with semantic segmentation. \textbf{\datasetname is the first semantically segmented LiDAR dataset collected with robotic systems in agricultural environments.}}
  
  \label{fig:platforms}
  \vspace{-.5cm}
\end{figure}

In forestry research, \gls{tls} has been commonly deployed to capture detailed under-canopy scans, but \gls{tls} is a static platform with limited mobility and is costly in terms of equipment and time to deploy and process.
\cite{White2016Remote}. 
Recent works have focused on segmenting trees from \gls{als} data \cite{White2016Remote, weiser2022individual}, which are LiDARs mounted onto small aircraft for over-canopy data collection. However, these methods have limitations estimating the \gls{dbh} due to tree canopy obstruction~\cite{Cao2022TreeSI, hyyppa2020undercanopy}. 
Some works have utilized \gls{mls} platforms in urban environments or small RC vehicles in agricultural settings \cite{Proudman2021Online}, but these techniques do not scale well as ground vehicles, which have limited mobility and efficiency when traversing through a cluttered forest on possibly undulating terrain \cite{Disney2018Weighing}.

Recently, autonomous UAVs have been able to fly reliably under-canopy~\cite{liu2021large, Loquercio2021Science}, enabling fast collection of under-canopy \gls{uls} data. Despite recent innovations from the robotics industry (TreeSwift, Gaia AI) and researchers 
~\cite{Proudman2021Online, Tremblay2019Automatic} in estimating biomass volume from mobile platforms, there is minimal public data or existing benchmarks to evaluate these algorithms. 

\datasetname aims to address this gap with the following contributions:
\begin{itemize}
    \item \textbf{We present the first LiDAR dataset of segmented tree stem point clouds acquired from mobile robots in forestry and orchard environments}.
    \item We provide manually annotated \textbf{semantic labels for the tree stems, and ground-truth field measurements of tree diameters}. 
    \item We share \textbf{benchmark scripts to evaluate diameter estimation algorithm and semantic segmentation performance} compared with the provided ground-truth data, and \textbf{share our baseline results with our novel diameter estimation algorithms} ~\cite{chen2019sloam} and \textbf{state-of-the-art segmentation networks} \cite{Milioto2019RangeNet, wu2018squeezesegv2, xu2020squeezesegv3}.
\end{itemize}

Our goal is for this dataset to enable an accurate evaluation of tree diameter estimation in robotic applications and the development of point cloud segmentation algorithms for agriculture by providing measurable performance metrics.


\section{Related Work}

The availability of annotated 3D LiDAR datasets \cite{behley2019semantickitti, pan2020semanticposs} \cite{varney2020dales} have accelerated the development of deep learning techniques for semantic segmentation for autonomous driving applications \cite{gao2020hungry}, but these datasets are exclusively collected in urban environments. 
Lately, there has been increased interest among forestry researchers for LiDAR-based data \cite{weiser2022terrestrial, burt2019extracting} to develop deep learning techniques for semantic segmentation and diameter estimation for agriculture. 

\subsection{LiDAR Semantic Segmentation}
Recent works have focused on using \gls{tls} to perform high-quality 3D reconstructions of forests for trunk and crown semantic segmentation \cite{Krisanski2021Sensor, burt2019extracting}.
Krisanski et al. \cite{Krisanski2021Sensor} present deep learning methods to accurately segment terrain, vegetation, and stems, and evaluate them in a public \gls{tls} dataset~\cite{Calders2014Terrestrial} and their own \gls{als}, \gls{uls}, and \gls{mls} data.
{Burt et al. \cite{burt2019extracting} use geometric methods and unsupervised learning to perform trunk and crown segmentation on larger-area forestry point clouds gathered from \gls{tls} data.

However, most forestry segmentation techniques require high-quality TLS point clouds and are rarely transferable to noisier clouds acquired from other platforms \cite{Krisanski2021Sensor}. 
Traditional \gls{tls} systems are impaired by the occlusion problem \cite{hyyppa2020undercanopy}, which hinders comprehensive tree scans. 
Furthermore, TLS is difficult to collect at scale in complex environments such as forests due to its limited mobility and efficiency ~\cite{Disney2018Weighing}. 
Notably, TLS is only feasible at the same scale as manual field-based survey measurements, as a single hectare plot takes 3 to 6 person days to scan \cite{Disney2018Weighing}. 
TLS datasets are also static, meaning they lack temporal context and spatiotemporal features between frames that are advantageous for deep learning models when trained with sequential datasets \cite{gao2020hungry}.

Additional works have demonstrated interest in decoupling from over-reliance on traditional TLS platforms, by developing algorithms compatible with data acquired from aerial and mobile platforms. 
Weiser et al. ~\cite{weiser2022terrestrial} presented forestry data collected from heterogeneous platforms for tree segmentation. They applied unsupervised learning and geometric methods to segment \gls{tls} data and manual annotations to the \gls{als} and \gls{uls} point clouds ~\cite{weiser2022individual}.
Tatsumi et al. \cite{Tatsumi2023ForestScanner} present methods for mapping trees with LiDAR-equipped iPhones to enable time-efficient forest inventories. 

\subsection{Tree Diameter Estimation}

Recent works have leveraged \gls{mls} data collection to enable novel diameter estimation algorithms. Tremblay et al. \cite{Tremblay2019Automatic} deployed mobile robots \gls{mls} to measure tree diameters in forests automatically and compared them to 943 ground-truth measurements.
Proudman et al. \cite{Proudman2021Online} presents a handheld \gls{mls} system that performs segmentation and \gls{dbh} estimation in real-time. 
Outside of the work done by Liu et al. ~\cite{liu2021large} and Chen et al. ~\cite{chen2019sloam} from our research group, Hyyppa et al. \cite{hyyppa2020undercanopy} is the only robotics study to use under-canopy ULS manual flight data, for \gls{dbh} and stem volume estimation on a data sample of 85 trees over 0.2 ha of forest plots.

However, to our knowledge there are very few datasets for laser scanning and robotics in forestry, especially with under-canopy data \cite{nunes2022procedural}.
Cao et al. \cite{Cao2022TreeSI} found that instance segmentation algorithms performed poorly for understory trees, and suggests that we need better under-canopy data to improve segmentation techniques. 
Due to the difficulty of obtaining annotated data from agricultural environments, some works are focusing on generating synthetic forestry datasets for robotic perception and machine learning development \cite{nunes2022procedural}. 
Neuville et al. \cite{Neuville2021Estimating} note that the over-canopy \gls{uls} has limitations in segmenting tree trunks and suggests that UAVs that can both operate above and under the canopy are a promising solution to gather data to tackle this issue.

To our knowledge, our work is the \textbf{first dataset of 3D LiDAR segmentation in agricultural environments gathered using robots, with our novel \gls{uls} and \gls{mls} mobile platforms} such as the Falcon 4 UAV and versatile \MakeLowercase{\sensortower}. 
Furthermore, we provide a comprehensive repository of \textbf{field-measured ground-truth tree diameters from diverse environments and corresponding individual tree point clouds from heterogeneous platforms to enable the development of diameter and biomass volume estimation algorithms}.
A summary of the differences between our work and existing literature is shown in Table \ref{tab:lidar_datasets}.

\begin{table*}[!htbp]
\footnotesize
\setlength{\tabcolsep}{4pt}
\centering
\begin{tabular}{|c|c|c|c|c|c|c|c|}
\hline
\textbf{Dataset} & \textbf{Platform} & \textbf{Size} & \textbf{Terrain} & \textbf{LiDAR} & \textbf{Ground Truth} & \textbf{Semantic Labels} & \textbf{Size}\\ \hline

RUSH06  & TLS & 0.5 ha & Eucalypt Open Forest & RIEGL VZ-400 & N/A & Tree cylinder & 22 GB \\ 
RUSH07 \cite{Calders2014Terrestrial} & &  & &  & & models & \\ \hline
NOU-11 \cite{burt2019extracting} & TLS & 1 ha & French Guiana, Cayenne & RIEGL VZ-400 & N/A & N/A & 64 GB \\ \hline 
KARA-001 \cite{burt2019extracting} & TLS & 0.25 ha & Karawatha Forest Park & RIEGL VZ-400 & N/A & N/A & N/A \\ \hline
BR01-08 & ALS & 12 ha & Bretten and & ALS: RIEGL VQ-780i   & DBH for 1060  & Single-tree & 105 GB \\ 
KA09-12 \cite{weiser2022terrestrial} & ULS & & Hardtwald forest & ULS: miniVUX-1UAV & ALS/ULS for 1491  & point clouds & \\ 
& TLS & & & TLS: RIEGL VZ-400 & TLS for 249  & (includes canopy) & \\ \hline
 WY-01$^*$ & TLS & 2 ha & Temperate broadleaf and  &   TLS: RIEGL VZ-400 & N/A & Trees & N/A \\ 
 SE-01$^*$ \cite{Cao2022TreeSI} & ALS &   & Tropical rainforest & ALS: Not Specified & & segmented & \\  \hline
 Hyyppä et al.$^*$  & ULS & 0.2 ha & Boreal forest in Finland  &   ULS: Velodyne VLP-16 & N/A & N/A & N/A \\ 
 \cite{hyyppa2020undercanopy} &  & & pine, spruce, birch &  & & & \\ \hline
 \textbf{\datasetname} & ULS & \textbf{4 ha of} & Pine, oak, maple, & ULS: Ouster OS1-64 & \textbf{DBH for 1860} & \textbf{HDF5 labels} & \textbf{2.2 TB} \\
 & MLS & \textbf{forests} & and cedar forests; & MLS: Ouster OS0-128 & \textbf{MLS for 1860} &  \textbf{for 3 classes} & \\ 
& & \textbf{46 ha of} & Almond and & and Velodyne VLP-16 & \textbf{ULS for 167} &  &  \\ 
&  & \textbf{orchards} & pistachio orchards &  & \textbf{Height profiles for 97} & &   \\ \hline
\end{tabular}
\caption{Comparison of LiDAR datasets used for tree semantic and instance segmentation. \datasetname is the only publicly available dataset acquired from robotics platforms that provides high-resolution LiDAR data, ground-truth DBH measurements, diameter-height profiles, and annotated semantic labels from heterogeneous agricultural environments. 
$^*$\emph{Dataset not publicly available}.} 
\label{tab:lidar_datasets}
\vspace{-.5cm}
\end{table*}



\section{Methods}
\datasetname provides sensor data and ROS bags collected with robotics platforms from agricultural environments, and manually annotated and field-measured ground-truth data.

\subsection{Sensor Stack}
\label{sec:sensor-stack}

We collected the data in this paper using two mobile platforms with similar sensor configurations: 
\begin{itemize}
    \item Falcon 4 UAV: UAV laser scanning (ULS) data was acquired in both manual teleoperated and autonomous flight modes.
    This platform has a total weight of 4.2 kg and can perform flights under-canopy for up to 30 minutes of flight ~\cite{liu2021large}. 
    The UAV is equipped with an Ouster OS1-64 LiDAR (Rev 6), with 64 vertical channels, 1024 horizontal points, 120 m range, and $45^{\circ}$ vertical FoV, and an Open Vision Computer~\cite{quigley2019ovc}.
    \item Our sensor platform: Mobile laser scanning (MLS) data was acquired with the \MakeLowercase{\sensortower}, which allows for a platform-agnostic approach to collecting robotics data, as it can be mounted on top of vehicles, small carts, carried on a backpack by a walking person, or even flown on small aircraft. 
    This features an Ouster OS0-128 LiDAR (Rev 6), with 128 vertical channels, 1024 or 2048 horizontal points, 50 m range, and $90^{\circ}$ vertical FoV, and an Intel RealSense D435 RGBD Camera.
\end{itemize}
Both platforms featured an Intel NUC10i7FNH onboard CPU, VectorNav VN-100 IMU, and a UBlox ZED-F9P GPS. An overview of the platforms used for data acquisition is shown in Figure~\ref{fig:platforms}.
Unless noted, only the LiDAR and IMU data are released in the dataset (except for the raw  data). 

\subsection{Data Collection}
Data was collected from a variety of forests and orchards: 
\begin{itemize}
  \item {Under-canopy \gls{uls} in Virginia and New Jersey state forests}.
  \item {ULS in Central California almond and pistachio orchards during canopy-on and canopy-off conditions}.
  \item {\gls{mls} data taken from our sensor tower platform (see Figure \ref{fig:platforms}) in forestry and orchard environments}.
  \item {Autonomous flight UAV sensor data for one hour of flight in the Appomattox-Buckingham State Forest}.
\end{itemize}

The tree species distribution for forestry datasets collected in Virginia (\textit{VAT-0723, VAT-1022}) consisted of loblolly (\textit{Pinus taeda}), Virginia pine (\textit{Pinus virginiana}), pitch pine (\textit{Pinus rigida}), eastern white pine (\textit{Pinus strobus}), chestnut oak (\textit{Quercus montana}), and white oak (\textit{Quercus alba}).
The data collected in New Jersey's Wharton State Forest (WSF-19) consisted of pitch pine (\textit{Pinus rigida}), various oak, and Atlantic white cedar (\textit{Chamaecyparis thyoides}).
The orchard datasets (\textit{UCM-0523, UCM-0323, UCM-0822}) gathered in Central California consist of almond orchards (\textit{Prunus dulcis}) and pistachio plantations (\textit{Pistacia vera}). 
Table \ref{tab:dataset_info} summarizes each of the datasets contained in \datasetname by providing details of the location, tree types, and type of collection, and Figure \ref{fig:diam-box-whisker} shows the \gls{dbh} distributions for each dataset's field measurements. 

The forestry data from the Appomattox-Buckingham State Forest (\textit{VAT-0723}) consists of 4 intensively managed plots (IMP), which are subject to manual thinning after 10-15 years to study the effects of crowding and competition factors on tree growth through metrics such as \gls{dbh}, crown height, and crown width. Lightly thinned (LT) plots cover 0.25 acres and have 50\% trees removed, and control plots (CT) cover 0.15 acres with no trees removed.

In some datasets, data is available from both platforms. A trailing letter is added to subset name to differentiate the LiDAR resolutions. (e.g., \textit{VAT-0723M} for MLS).

\subsection{Input Bags}
\label{input-bags}
The input ROS bags contain raw data from all onboard sensors as described in Sec. \ref{sec:sensor-stack}, including LiDAR, IMU, RGBD image, and GPS data. 
Processed ROS bags are also provided, which contain ground-truth lidar-intertial odometry and velocity-corrected point cloud frames for every sweep of the LiDAR provided by Faster-LIO \cite{Bai2022Faster}. 
We also provide semantically inferred ground and tree point clouds in the sensor and robot base frame, from our trained segmentation network based on the RangeNet++ \cite{Milioto2019RangeNet} architecture. In total, there are over ten hours of raw data from the ROS bags.

\begin{table*}[h]
  \setlength{\tabcolsep}{4pt}
  \centering
  \begin{tabular}{|c|c|c|c|c|c|c|}
    \hline
    \textbf{Dataset} & \textbf{Location} & \textbf{Size} & \textbf{Trees} & \textbf{LiDAR} & \textbf{Labeled Frames} & \textbf{GT DBH}  \\[1mm]
    \hline
    VAT-0723U & Appomattox-Buckingham & CT: 0.12 ha & loblolly pine & OS1-64 (ULS) & 183 & $77^\dagger$   \\
    VAT-0723M & State Forest, VA & LT: 0.2 ha & & OS0-128 (MLS) & 145 &    \\ [1mm]
    \hline
    VAT-1022 & Fishburn Forest, VA & 0.5 ha & Virginia, pitch, \& eastern white pine & OS1-64 (ULS) & 201 & $20^{\dagger\ddagger}$ \\
     &  &  & White \& chestnut oak &  &  &    \\ [1mm]
    \hline 
    WSF-19 & Wharton State  & 3 ha & pitch pine, various oak,  & VLP-16 (MLS) & 52 & 1,539\\
     & Forest, NJ  &  & Atlantic white cedar & & &  \\ [1mm]
    \hline 
    UCM-0523U & Merced County, CA & 1 ha & pistachio orchard & OS1-64 (ULS) & 400 & 70$^*$ \\
    UCM-0523M & & & & OS0-128 (MLS) & 434 & \\
    \hline
    UCM-0323 & Central CA & 25 ha & pistachio orchard & OS0-128 (MLS) & 242 & 154*  \\[1mm]
    \hline
    UCM-0822 & Merced County, CA & 20 ha & almond orchard & OS0-128 (MLS) & 187 & N/A \\[1mm]
    \hline
  \end{tabular}
  \caption{An overview of the data provided by \datasetname: ground-truth with manually annotated semantic labels and field-measured \gls{dbh} from LiDAR data acquired from UAV laser scanning (ULS) and mobile laser scanning (MLS) platforms. \\
  $^\dagger$\emph{Tree heights provided}. \\
  $^\ddagger$\emph{Full diameter profiles are provided at 1 m height intervals to enable ground-truth volume benchmark}. \\
  $^*$\emph{Diameters measured at 0.3 m height for short orchard trees}.
  }
  \label{tab:dataset_info}
\end{table*}

\subsection{Semantic Labels}
\label{semantic-labels}
Manually annotated semantic labels for tree stems, ground points, and miscellaneous are provided for each data subset and model resolution platform.
We developed a labeling tool that aggregates and undistorts LiDAR frames based on LIO mapping ~\cite{Bai2022Faster, llol2022qu}, allowing us to accrue multiple scans within a single labeling process.
The accumulated point cloud is organized using a spherical projection model to project it into the 2-D range image space. 
Then, it uses a top-down labeling approach where scans are indexed based on the z-axis height of capture, allowing for hierarchical labeling of multi-structured data. 
The labeling process is shown in Figure \ref{fig:sill-labeling}.


\begin{figure}[b]
    \centering
    \includegraphics[width=\linewidth]{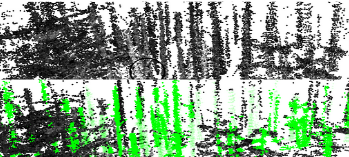}
    \caption{Manual semantic annotations using our labeling tool.  \\ \ul{Top:} 40 accumulated raw scans of the Virginia Tech forest. \\ \ul{Bottom:} Tree stems labeled in green. (\textit{VAT-1022} dataset)} 
    \label{fig:sill-labeling}
    \vspace{-.5cm}
\end{figure}


\subsection{Ground-Truth Trees}
Hand-measured ground-truth diameters are provided for trees in each dataset as described in Table \ref{tab:dataset_info}. The diameters are measured at the standard \gls{dbh} at 4.5 ft (1.37 m) above ground, and the \gls{dbh} distributions are shown in Figure \ref{fig:diam-box-whisker}. Furthermore, measured tree heights are provided for 97 forestry trees, and full diameter profiles at 1 meter height intervals are provided for all 20 trees in the \textit{VAT-1022} forestry dataset. For each tree with a ground-truth measurement, individual tree point clouds are provided, as well as diameter estimations from our novel methods as described in Sec. \ref{dbh-methods}. 

In the \textit{WSF-19} dataset, we measured the variance in field measurements between two humans when measuring DBH of the same tree. Across 695 trees evaluated, there was an RMSE of 0.31 cm between human-to-human measurements, equal to 1.5\% of the DBH in the sample size, indicating the field measurements have minimal variance across different humans.

\begin{figure}[b]
    \centering
    \includegraphics[width=0.9\linewidth]{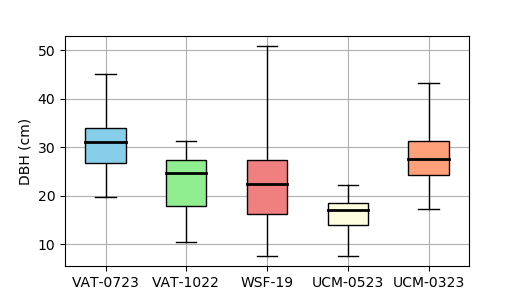}
    \caption{Ground-truth diameter distributions for each dataset. Most forestry trees (\textit{VAT-0723, VAT-1022, WSF-19}) have similar \gls{dbh} values, but due to the large sample size, there is a wide distribution of min/max \gls{dbh}. Orchard trees (\textit{UCM-0523, UCM-0323}) are planted in organized environments and have less variance in \gls{dbh}. 
    }
    \label{fig:diam-box-whisker}
    \vspace{-.5cm}
\end{figure}

\section{Dataset Applications and Metrics}
\label{sec:metrics}
\subsection{Semantic Segmentation}



 The annotated point cloud labels can be used to train a segmentation network to provide semantic and instance segmentation on tree stem and ground points. We use Intersection over Union (IoU) \cite{Hurtado2022Semantic} as the performance metric for semantic segmentation results. The per-class IoU equation is given below in Equation \ref{eqn:iou}, where $c_{ii}$,  $c_{ij}$, and $c_{ki}$ denote truly positive, false positive, and false negative predictions respectively.
\begin{align}
\label{eqn:iou}
IoU_i = \frac{c_{ii}}{c_{ii} + \sum_{j \neq i} c_{ij} + \sum_{k \neq i} c_{ki}}
\end{align}

Overall accuracy (OA) \cite{Hurtado2022Semantic} is also used as a standard benchmark in point cloud datasets. Note that the overall accuracy is often skewed by the imbalanced distribution of points across classes (see class distributions in Table \ref{tab:segmentation_baseline}). 
\begin{align}
\label{eqn:oa}
OA = \frac{\sum_{i=1}^{N} c_{ii}}{N \sum_{j=1}^{N} \sum_{k=1}^{N} c_{jk}}
\end{align}
 

\begin{figure}[t]
    \centering
    \includegraphics[width=\linewidth]{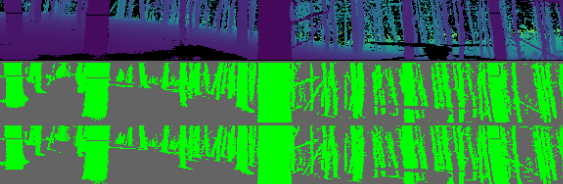}
    \caption{Segmentation results from the RangeNet++ ~\cite{Milioto2019RangeNet} model. The inference mask (bottom) is compared on a pixel-by-pixel basis to the ground-truth mask (middle) across the test set to compute the segmentation performance metrics. \\ \ul{Top:} Range image 2D projection of input LiDAR scan. \ul{Middle:} Ground-truth semantic labels (green pixels are tree stems). \ul{Bottom:} Segmentation inference from the model.}
    \label{fig:fig4}
\end{figure}

\subsection{Diameter Estimation} 
The input bags for this dataset, along with the individual tree point clouds and corresponding DBH ground-truth measurements, can be used to test the accuracy of diameter estimation algorithms across a diverse set of trees and environments. 
The root-mean-square error (RMSE) of the diameter estimation ($\hat{d_i}$) compared to ground-truth data ($d_i$) is used to evaluate the performance of diameter estimation algorithms. 
\begin{equation}
\label{eqn:rmse}
RMSE = \sqrt{\frac{1}{N} \sum_{i=1}^{N} (d_i - \hat{d}_i)^2}
\end{equation}

\section{Evaluation}
We share benchmark scripts\footnote{Benchmark scripts are available at \url{https://github.com/KumarRobotics/treescope/main/benchmarks}.} to help researchers evaluate their methods using the metrics described in Sec. \ref{sec:metrics}, and provide examples of algorithms and baseline results in this section.

\subsection{Semantic Segmentation}

We evaluate three start-of-the-art segmentation \cite{Milioto2019RangeNet, wu2018squeezesegv2, xu2020squeezesegv3} networks with the available open-source code from the original authors on various data subsets of \datasetname and report the baseline results in Table \ref{tab:segmentation_baseline}. 
RangeNet++ \cite{Milioto2019RangeNet} outperforms SqueezeSegV3 \cite{xu2020squeezesegv3} across most categories in terms of OA and IoU metrics. However, SqueezeSegV2 \cite{wu2018squeezesegv2} has a significantly faster inference speed. For real-time inference on high-resolution LiDARs such as the 128x2048 Ouster OS0-128 on our \MakeLowercase{\sensortower} \gls{mls} platform, a lightweight method like SqueezeSegV2 may be preferable for its faster inference time despite its lower segmentation performance. 



\begin{table*}[h]
  \centering
  \begin{tabular}{|c|c|c|c|ccc|ccc|c|}
    \hline
     &  &  & &   \multicolumn{3}{c|}{IoU} & \multicolumn{3}{c|}{Class Distribution} & Inference Time \\
     & Dataset \& Size & Method & OA & Mean & Tree Stem & Ground & Tree Stem & Ground & Misc & [ms/Scan]\\
    \hline  
    \multirow{6}{*}{\STAB{\rotatebox[origin=c]{90}{Forests}}}
     & VAT-0723U & \textbf{RangeNet53++} \cite{Milioto2019RangeNet} & \textbf{0.875} & \textbf{0.662} & \textbf{0.684} & {0.451} & 0.245 & 0.096 &  0.659 & 47 \\
     & 64x1024 px & SqueezeSegV2 \cite{wu2018squeezesegv2} & {0.615} & {0.429} & {0.451} & {0.395} & 0.245 & 0.096 &  0.659 & \textbf{12} \\
     &  & SqueezeSegV3-53 \cite{xu2020squeezesegv3} &  {0.866} & {0.657} & {0.661} & \textbf{0.462} & 0.245 & 0.096 &  0.659 & 93 \\
     \cline{2-11}
     & VAT-1022 & \textbf{RangeNet53++} \cite{Milioto2019RangeNet} & \textbf{0.966} & \textbf{0.897} & \textbf{0.761} & \textbf{0.963} & 0.468 & 0.205 & 0.327 & 47 \\
     & 64x1024 px  & SqueezeSegV2 \cite{wu2018squeezesegv2} &  {0.832} & {0.762} & {0.526} & {0.835} & 0.468 & 0.205 & 0.327 & \textbf{12} \\
     & & SqueezeSegV3-53 \cite{xu2020squeezesegv3} &  {0.955} & {0.886} & {0.745} & {0.955} & 0.468 & 0.205 & 0.327 & 93 \\
     \hline
     \multirow{6}{*}{\STAB{\rotatebox[origin=c]{90}{Orchards}}}
    & UCM-0523M & \textbf{RangeNet53++} \cite{Milioto2019RangeNet} & \textbf{0.982} & \textbf{0.844} & \textbf{0.611} & \textbf{0.943} & 0.003 & 0.188 & 0.809 & 168 \\
    & 128x2048 px  & SqueezeSegV2 \cite{wu2018squeezesegv2} &  {0.913} & {0.730} & {0.407} & {0.835}  & 0.003 & 0.188 & 0.809 & \textbf{44} \\
     & & SqueezeSegV3-53 \cite{xu2020squeezesegv3} &  {0.975} & {0.837} & {0.599} & {0.938} & 0.003 & 0.188 & 0.809 & 321 \\
     \cline{2-11}
    & UCM-0323 & RangeNet53++ \cite{Milioto2019RangeNet} & 0.905 & 0.687 & 0.508 & \textbf{0.690} & 0.010 & 0.097 & 0.893 & 168 \\
     & 128x2048 px  & SqueezeSegV2 \cite{wu2018squeezesegv2} &  {0.692} & {0.561} & {0.238} & {0.527} & 0.010 & 0.097 & 0.893 & \textbf{44} \\
     & & \textbf{SqueezeSegV3-53} \cite{xu2020squeezesegv3} &  \textbf{0.912} & \textbf{0.692} & \textbf{0.523} & {0.682} & 0.010 & 0.097 & 0.893 & 321 \\
    \hline

    \hline
  \end{tabular}
  \caption{Segmentation results on \datasetname with off-the-shelf networks. We report the overall accuracy, mean IoU, and per-class IoU for each category. Note the datasets have a large variance in class distributions between object categories. RangeNet++ \cite{Milioto2019RangeNet} and SqueezeSegV3 \cite{xu2020squeezesegv3} were both deployed with a 53-layer backbone. RangeNet++ has better performance across most categories and data subsets. SqueezeSegV2 \cite{wu2018squeezesegv2} has the fastest inference time.}
  
  \label{tab:segmentation_baseline}
\end{table*}

\subsection{Diameter Estimation}
\label{dbh-methods}
We evaluate our two novel diameter estimation methods described in Sec.~\ref{sec:metrics} with our benchmarks and report the results in Table \ref{tab:dbh_baseline}.
\subsubsection{Density-based clustering ring extraction (DBCRE)} 
\label{dbh-dbscan}
DBCRE uses unsupervised clustering DBSCAN \cite{Ester1996DBSCAN} on offline accumulated point clouds to perform instance segmentation on the tree trunks.
First, $n$ sets of evenly spaced \emph{control points} on individual tree clusters are sampled, then a $k$-$d$ tree \cite{bentley1975kdtrees} is built out of the tree trunk instance to enable fast nearest neighbor searches. 
Using the properties of $k$-$d$ trees \cite{bentley1975kdtrees}, each set of $n$ \emph{control points} indexes a set of $m$ ring points that fall on the same height as the control points. The ring points are selected to correspond to the curvature of the tree trunks as seen from the LiDAR's perspective. 

The diameter calculations are done per-frame to counteract any odometry drift over time. The control points are accumulated in space and re-clustered using DBSCAN to associate them with the corresponding tree instances as clustered in the initial step. 
Finally, diameter is calculated from the largest chord formed among the set of ring points $\{{r_1, r_2, r_m}\}$.
Thus, each tree's diameter is $D_i = max(E(R_j))$, where $D_i$ is the diameter of the $i^{th}$ control point from the Euclidean distance $E$ indexed from $R_j$ ring points at its height. 

\subsubsection{Semantic lidar odometry and mapping (SLOAM) } \label{dbh-sloam}
SLOAM ~\cite{chen2019sloam} is a real-time state estimation and mapping software that models trees and ground points as cylinder and plane landmarks respectively, and estimates pose transform by performing data association with landmarks across current and previous LiDAR sweeps to register the LiDAR point clouds. Tree cylinder parameters are given by $s = (\rho, \phi, \nu, \alpha, \kappa)$, where $\rho$ is the distance between the origin and cylinder, $\phi$ and $\nu$ are the angles between the cylinder normal with the $x$ and $y$ axis respectively, $\alpha$ is the partial derivative of the cylinder normal with respect to $\nu$, and $\kappa$ is the inverse radius of the cylinder. Solving the geometric least squares optimization shown in Equation \ref{eqn:sloam} yields a cylinder tree model $s$ with diameter $2/\kappa$. The full methodology is described in Section III of Chen et al. ~\cite{chen2019sloam}.
\begin{equation}
\label{eqn:sloam}
\operatorname*{argmin}_{\rho,\phi,\nu,\alpha,\kappa}  \sum^{\delta_{i,k+1}}_{j=0} \hat{d}_s(s, p_j)
\end{equation}

As shown in Table \ref{tab:dbh_baseline}, DBCRE outperforms SLOAM across all datasets. SLOAM ~\cite{chen2019sloam} requires dense point clouds for accurate real-time clustering, so its accuracy was significantly worse with lower-resolution LiDAR. However, SLOAM has the advantage of being an online method that can generate cylinder landmarks and diameter estimations at 2 Hz, while DBCRE is an offline method that requires accumulation of points over the entire ROS bag.

\begin{table*}[h]
  \centering
  \begin{tabular}{|c|c|c|c|c|cc|cc|}
    \hline
    & Dataset & Platform & N & Mean DBH & DBCRE RMSE & RMSE \% & SLOAM RMSE & RMSE \% \\ 
    \hline
    \multirow{4}{*}{\STAB{\rotatebox[origin=c]{90}{Forests}}}
    & VAT-0723U & ULS & 77 & 30.8 & \textbf{3.8} & \textbf{12.3\%} &9.8 & 31.8\% \\ [0.7mm]
    & VAT-0723M & MLS & 77 & 30.8 & \textbf{3.5} & \textbf{11.6\%} &7.0 & 22.7\% \\ [0.7mm]
    & VAT-1022 & ULS & 20 & 27.3 & \textbf{2.0} & \textbf{7.5\%} &7.5 & 27.5\% \\ [0.7mm]
    & WSF-19 & MLS & 1539 & 23.1 & & & \textbf{2.3}  & \textbf{9.9\%} \\ [0.7mm]
    \hline
    \multirow{4}{*}{\STAB{\rotatebox[origin=c]{90}{Orchards}}}
    & UCM-0523U & ULS & 70 & 16.3 & \textbf{2.3} & \textbf{14.1\%} &5.1 & 31.3\% \\ [0.9mm]
    & UCM-0523M & MLS & 70 & 16.3 & \textbf{2.6} & \textbf{16.0\%} &3.8 & 23.4\% \\ [0.9mm]
    & UCM-0323 & MLS & 154 & 27.8 &\textbf{6.1} & \textbf{22.1\%} & 9.1 & 32.7\% \\ [0.9mm]
    
    \hline
  \end{tabular}
  \caption{Diameter estimation results compared to ground-truth (all values in cm) on \datasetname using our novel methods described in Sec. \ref{dbh-methods}. DBCRE outperforms SLOAM \cite{chen2019sloam} across all datasets, but SLOAM has the advantage of being an online method that we have also used for autonomous flight.}
  \label{tab:dbh_baseline}
\end{table*}

\section{Conclusions}
\subsection{Contributions}

This paper described \datasetname v1.0, a state-of-the-art LiDAR dataset for under-canopy forestry and orchard trees. Our goal is that \datasetname will allow researchers to develop solutions for agricultural applications by providing diverse data in challenging conditions that were not explored by other datasets previously. We encourage researchers to use this dataset to implement and test their semantic segmentation and diameter estimation methods. Overall, we hope that \datasetname will become a new standard for developing LiDAR-based algorithms for applications in agricultural robotics. 


\subsection{Data Availability}

The data will be fully open and available to researchers and educators for non-commercial use under the \textbf{Creative Commons Attribution-NonCommercial-ShareAlike 4.0 International License}\footnote{Creative Commons Attribution Non-Commercial 4.0 License: \url{https://creativecommons.org/licenses/by-nc-sa/4.0/}.}. Furthermore, the entire data stack will be available for download to allow development at any level without the need to re-create. 

We provide JSON files to describe ground-truth diameter and height profiles, dataset metadata, and LiDAR sensor metadata. We also provide HDF5 files with annotated semantic labels for tree stem, ground, and miscellaneous classes as described in Sec. \ref{semantic-labels}. As described in Sec. \ref{input-bags}, raw and processed ROS bags are provided. Finally, we share benchmark scripts to allow quick comparison of semantic segmentation and diameter estimation results against the provided ground-truth data.

\subsection{Further Work}
This is Version 1.0 of \datasetname, as we intend to publish further extensions of the work presented here. In the next iteration of this dataset, we aim to augment the data with canopy and branch semantic labels, which will enable robust timber biomass volume estimation algorithms. We may also release data from the other sensors available in our Falcon 4 UAV and custom sensor platforms, which are not included in this current release.




\section*{Acknowledgements}
We would like to thank Steven Chen for his contributions to prior data collection and processing of the \textit{WSF-19} data, Tom Donnelly for his preliminary data collection with the \textit{UCM-0822} data, Siming He for his work with the software and hardware stack of the sensor platform,  Ian Miller for his development of the Semantic Integrated LiDAR Labeling tool (SILL) used for ground-truth semantic annotations, Jeremy Wang for support with design and machining of parts, and Alex Zhou for designing and maintaining our UAVs.
\bibliographystyle{IEEEtran}
\bibliography{IEEEabrv, eg}

\begin{thebibliography}{10}
\providecommand{\url}[1]{#1}
\csname url@rmstyle\endcsname
\providecommand{\newblock}{\relax}
\providecommand{\bibinfo}[2]{#2}
\providecommand\BIBentrySTDinterwordspacing{\spaceskip=0pt\relax}
\providecommand\BIBentryALTinterwordstretchfactor{4}
\providecommand\BIBentryALTinterwordspacing{\spaceskip=\fontdimen2\font plus
\BIBentryALTinterwordstretchfactor\fontdimen3\font minus
  \fontdimen4\font\relax}
\providecommand\BIBforeignlanguage[2]{{%
\expandafter\ifx\csname l@#1\endcsname\relax
\typeout{** WARNING: IEEEtran.bst: No hyphenation pattern has been}%
\typeout{** loaded for the language `#1'. Using the pattern for}%
\typeout{** the default language instead.}%
\else
\language=\csname l@#1\endcsname
\fi
#2}}

\bibitem{brown1997estimating}
S.~Brown, \emph{Estimating biomass and biomass change of tropical forests: a
  primer}.\hskip 1em plus 0.5em minus 0.4em\relax Food \& Agriculture Org.,
  1997, vol. 134.

\bibitem{Calders2022Carbon}
\BIBentryALTinterwordspacing
K.~Calders, H.~Verbeeck, A.~Burt, N.~Origo, J.~Nightingale, Y.~Malhi,
  P.~Wilkes, P.~Raumonen, R.~G.~H. Bunce, and M.~Disney, ``Laser scanning
  reveals potential underestimation of biomass carbon in temperate forest,''
  \emph{Ecological Solutions and Evidence}, vol.~3, no.~4, p. e12197, 2022.
  [Online]. Available:
  \url{https://besjournals.onlinelibrary.wiley.com/doi/abs/10.1002/2688-8319.12197}
\BIBentrySTDinterwordspacing

\bibitem{Krisanski2021Sensor}
\BIBentryALTinterwordspacing
S.~Krisanski, M.~S. Taskhiri, S.~Gonzalez~Aracil, D.~Herries, and P.~Turner,
  ``Sensor agnostic semantic segmentation of structurally diverse and complex
  forest point clouds using deep learning,'' \emph{Remote Sensing}, vol.~13,
  no.~8, 2021. [Online]. Available:
  \url{https://www.mdpi.com/2072-4292/13/8/1413}
\BIBentrySTDinterwordspacing

\bibitem{White2016Remote}
\BIBentryALTinterwordspacing
J.~C. White, N.~C. Coops, M.~A. Wulder, M.~Vastaranta, T.~Hilker, and
  P.~Tompalski, ``Remote sensing technologies for enhancing forest inventories:
  A review,'' \emph{Canadian Journal of Remote Sensing}, vol.~42, no.~5, pp.
  619--641, 2016. [Online]. Available:
  \url{https://doi.org/10.1080/07038992.2016.1207484}
\BIBentrySTDinterwordspacing

\bibitem{liu2022challenges}
X.~Liu, S.~W. Chen, G.~V. Nardari, C.~Qu, F.~C. Ojeda, C.~J. Taylor, and
  V.~Kumar, ``Challenges and opportunities for autonomous micro-uavs in
  precision agriculture,'' \emph{IEEE Micro}, vol.~42, no.~1, pp. 61--68, 2022.

\bibitem{liu2021large}
X.~Liu, G.~V. Nardari, F.~C. Ojeda, Y.~Tao, A.~Zhou, T.~Donnelly, C.~Qu, S.~W.
  Chen, R.~A.~F. Romero, C.~J. Taylor, and V.~Kumar, ``{Large-Scale Autonomous
  Flight With Real-Time Semantic SLAM Under Dense Forest Canopy},'' \emph{IEEE
  Robotics and Automation Letters}, vol.~7, no.~2, pp. 5512--5519, 2022.

\bibitem{weiser2022individual}
\BIBentryALTinterwordspacing
H.~Weiser, J.~Sch\"afer, L.~Winiwarter, N.~Kra\v{s}ovec, F.~E. Fassnacht, and
  B.~H\"ofle, ``Individual tree point clouds and tree measurements from
  multi-platform laser scanning in german forests,'' \emph{Earth System Science
  Data}, vol.~14, no.~7, pp. 2989--3012, 2022. [Online]. Available:
  \url{https://essd.copernicus.org/articles/14/2989/2022/}
\BIBentrySTDinterwordspacing

\bibitem{Cao2022TreeSI}
Y.~Cao, J.~G.~C. Ball, D.~A. Coomes, L.~Steinmeier, N.~Knapp, P.~Wilkes, M.~I.
  Disney, K.~Calders, A.~Burt, Y.~Lin, and T.~Jackson, ``Tree segmentation in
  airborne laser scanning data is only accurate for canopy trees,''
  \emph{bioRxiv}, 2022.

\bibitem{hyyppa2020undercanopy}
\BIBentryALTinterwordspacing
E.~Hyyppä, J.~Hyyppä, T.~Hakala, A.~Kukko, M.~A. Wulder, J.~C. White,
  J.~Pyörälä, X.~Yu, Y.~Wang, J.-P. Virtanen, O.~Pohjavirta, X.~Liang,
  M.~Holopainen, and H.~Kaartinen, ``Under-canopy uav laser scanning for
  accurate forest field measurements,'' \emph{ISPRS Journal of Photogrammetry
  and Remote Sensing}, vol. 164, pp. 41--60, 2020. [Online]. Available:
  \url{https://www.sciencedirect.com/science/article/pii/S0924271620300915}
\BIBentrySTDinterwordspacing

\bibitem{Proudman2021Online}
A.~Proudman, M.~Ramezani, and M.~Fallon, ``Online estimation of diameter at
  breast height (dbh) of forest trees using a handheld lidar,'' in \emph{2021
  European Conference on Mobile Robots (ECMR)}, 2021, pp. 1--7.

\bibitem{Disney2018Weighing}
M.~Disney, M.~Boni~Vicari, A.~Burt, K.~Calders, S.~Lewis, P.~Raumonen, and
  P.~Wilkes, ``Weighing trees with lasers: Advances, challenges and
  opportunities,'' \emph{Interface Focus}, vol.~8, p. 20170048, 04 2018.

\bibitem{Loquercio2021Science}
A.~Loquercio, E.~Kaufmann, R.~Ranftl, M.~M{\"u}ller, V.~Koltun, and
  D.~Scaramuzza, ``Learning high-speed flight in the wild,'' in \emph{Science
  Robotics}, October 2021.

\bibitem{Tremblay2019Automatic}
\BIBentryALTinterwordspacing
J.-F. Tremblay, M.~Béland, F.~Pomerleau, R.~Gagnon, and P.~Giguère,
  ``Automatic 3d mapping for tree diameter measurements in inventory
  operations,'' 2019. [Online]. Available:
  \url{https://arxiv.org/abs/1904.05281}
\BIBentrySTDinterwordspacing

\bibitem{chen2019sloam}
S.~W. Chen, G.~V. Nardari, E.~S. Lee, C.~Qu, X.~Liu, R.~A.~F. Romero, and
  V.~Kumar, ``Sloam: Semantic lidar odometry and mapping for forest
  inventory,'' in \emph{IEEE Robotics and Automation Letters (RA-L)}, 2020.

\bibitem{Milioto2019RangeNet}
A.~Milioto, I.~Vizzo, J.~Behley, and C.~Stachniss, ``Rangenet ++: Fast and
  accurate lidar semantic segmentation,'' in \emph{2019 IEEE/RSJ International
  Conference on Intelligent Robots and Systems (IROS)}, 2019, pp. 4213--4220.

\bibitem{wu2018squeezesegv2}
B.~Wu, X.~Zhou, S.~Zhao, X.~Yue, and K.~Keutzer, ``Squeezesegv2: Improved model
  structure and unsupervised domain adaptation for road-object segmentation
  from a lidar point cloud,'' 2018.

\bibitem{xu2020squeezesegv3}
C.~Xu, B.~Wu, Z.~Wang, W.~Zhan, P.~Vajda, K.~Keutzer, and M.~Tomizuka,
  ``Squeezesegv3: Spatially-adaptive convolution for efficient point-cloud
  segmentation,'' in \emph{European Conference on Computer Vision}.\hskip 1em
  plus 0.5em minus 0.4em\relax Springer, 2020, pp. 1--19.

\bibitem{behley2019semantickitti}
J.~Behley, M.~Garbade, A.~Milioto, J.~Quenzel, S.~Behnke, C.~Stachniss, and
  J.~Gall, ``Semantickitti: A dataset for semantic scene understanding of lidar
  sequences,'' 2019.

\bibitem{pan2020semanticposs}
Y.~Pan, B.~Gao, J.~Mei, S.~Geng, C.~Li, and H.~Zhao, ``Semanticposs: A point
  cloud dataset with large quantity of dynamic instances,'' 2020.

\bibitem{varney2020dales}
N.~Varney, V.~K. Asari, and Q.~Graehling, ``Dales: A large-scale aerial lidar
  data set for semantic segmentation,'' 2020.

\bibitem{gao2020hungry}
B.~Gao, Y.~Pan, C.~Li, S.~Geng, and H.~Zhao, ``Are we hungry for 3d lidar data
  for semantic segmentation? a survey and experimental study,'' 2020.

\bibitem{weiser2022terrestrial}
H.~Weiser, J.~Sch{\"a}fer, L.~Winiwarter, N.~Kra{\v{s}}ovec, C.~Seitz,
  M.~Schimka, K.~Anders, D.~Baete, A.~Braz, J.~Brand, \emph{et~al.},
  ``Terrestrial, uav-borne, and airborne laser scanning point clouds of central
  european forest plots, germany, with extracted individual trees and manual
  forest inventory measurements,'' 2022.

\bibitem{burt2019extracting}
A.~Burt, M.~Disney, and K.~Calders, ``Extracting individual trees from lidar
  point clouds using treeseg,'' \emph{Methods in Ecology and Evolution},
  vol.~10, no.~3, pp. 438--445, 2019.

\bibitem{Calders2014Terrestrial}
K.~Calders, ``Terrestrial laser scans - riegl vz400, individual tree point
  clouds and cylinder models, rushworth forest. version 1. terrestrial
  ecosystem research network.'' https://doi.org/10.4227/05/542B766D5D00D, 2014.

\bibitem{Tatsumi2023ForestScanner}
\BIBentryALTinterwordspacing
S.~Tatsumi, K.~Yamaguchi, and N.~Furuya, ``Forestscanner: A mobile application
  for measuring and mapping trees with lidar-equipped iphone and ipad,''
  \emph{Methods in Ecology and Evolution}, vol.~14, no.~7, pp. 1603--1609,
  2023. [Online]. Available:
  \url{https://besjournals.onlinelibrary.wiley.com/doi/abs/10.1111/2041-210X.13900}
\BIBentrySTDinterwordspacing

\bibitem{nunes2022procedural}
\BIBentryALTinterwordspacing
R.~Nunes, J.~F. Ferreira, and P.~Peixoto, ``Procedural generation of synthetic
  forest environments to train machine learning algorithms,'' in \emph{ICRA
  2022 Workshop in Innovation in Forestry Robotics: Research and Industry
  Adoption}, 2022. [Online]. Available:
  \url{https://openreview.net/forum?id=rpzgjNCe4G9}
\BIBentrySTDinterwordspacing

\bibitem{Neuville2021Estimating}
\BIBentryALTinterwordspacing
R.~Neuville, J.~S. Bates, and F.~Jonard, ``Estimating forest structure from
  uav-mounted lidar point cloud using machine learning,'' \emph{Remote
  Sensing}, vol.~13, no.~3, 2021. [Online]. Available:
  \url{https://www.mdpi.com/2072-4292/13/3/352}
\BIBentrySTDinterwordspacing

\bibitem{quigley2019ovc}
M.~Quigley, K.~Mohta, S.~S. Shivakumar, M.~Watterson, Y.~Mulgaonkar,
  M.~Arguedas, K.~Sun, S.~Liu, B.~Pfrommer, V.~Kumar, and C.~J. Taylor, ``{The
  Open Vision Computer: An Integrated Sensing and Compute System for Mobile
  Robots},'' in \emph{2019 International Conference on Robotics and Automation
  (ICRA)}, 2019, pp. 1834--1840.

\bibitem{Bai2022Faster}
C.~Bai, T.~Xiao, Y.~Chen, H.~Wang, F.~Zhang, and X.~Gao, ``Faster-lio:
  Lightweight tightly coupled lidar-inertial odometry using parallel sparse
  incremental voxels,'' \emph{IEEE Robotics and Automation Letters}, vol.~7,
  no.~2, pp. 4861--4868, 2022.

\bibitem{llol2022qu}
C.~Qu, S.~S. Shivakumar, W.~Liu, and C.~J. Taylor, ``Llol: Low-latency odometry
  for spinning lidars,'' in \emph{"2022 International Conference on Robotics
  and Automation (ICRA)"}, ser. 2022 IEEE International Conference on Robotics
  and Automation (ICRA).\hskip 1em plus 0.5em minus 0.4em\relax IEEE,
  2022-5-23.

\bibitem{Hurtado2022Semantic}
\BIBentryALTinterwordspacing
J.~V. Hurtado and A.~Valada, ``Chapter 12 - semantic scene segmentation for
  robotics,'' in \emph{Deep Learning for Robot Perception and Cognition},
  A.~Iosifidis and A.~Tefas, Eds.\hskip 1em plus 0.5em minus 0.4em\relax
  Academic Press, 2022, pp. 279--311. [Online]. Available:
  \url{https://www.sciencedirect.com/science/article/pii/B9780323857871000178}
\BIBentrySTDinterwordspacing

\bibitem{Ester1996DBSCAN}
M.~Ester, H.-P. Kriegel, J.~Sander, and X.~Xu, ``A density-based algorithm for
  discovering clusters in large spatial databases with noise,'' p. 226–231,
  1996.

\bibitem{bentley1975kdtrees}
\BIBentryALTinterwordspacing
J.~L. Bentley, ``Multidimensional binary search trees used for associative
  searching,'' \emph{Commun. ACM}, vol.~18, no.~9, p. 509–517, sep 1975.
  [Online]. Available: \url{https://doi.org/10.1145/361002.361007}
\BIBentrySTDinterwordspacing

\end{thebibliography}

\end{document}